%% file: main.tex
\documentclass[10pt,twocolumn,letterpaper]{article}

\usepackage[pagenumbers]{wacv}

\usepackage{graphicx}
\usepackage{amsmath}
\usepackage{amssymb}
\usepackage{booktabs}
\usepackage{blindtext}
\usepackage{comment}
\usepackage{xspace}
\usepackage{colortbl}
\usepackage{bm}
\usepackage{bbding}

\usepackage{pifont}
\newcommand{\cmark}{\ding{51}}
\newcommand{\xmark}{\ding{55}}
\usepackage[dvipsnames]{xcolor}
\usepackage[pagebackref,breaklinks,colorlinks]{hyperref}

\usepackage[capitalize]{cleveref}
\crefname{section}{Sec.}{Secs.}
\Crefname{section}{Section}{Sections}
\Crefname{table}{Table}{Tables}
\crefname{table}{Tab.}{Tabs.}

\def\wacvPaperID{1536}
\def\confName{WACV}
\def\confYear{2025}

\definecolor{mentor-green}{RGB}{0,170,70}
\definecolor{mentor-blue}{RGB}{60,100,190}

\begin{document}

\title{MENTOR:\\Human Perception-Guided Pretraining for Increased Generalization}

\author{Colton R. Crum, Adam Czajka\\ 
University of Notre Dame\\
384 Fitzpatrick Hall of Engineering, Notre Dame, IN 46556\\
{\tt\small ccrum@nd.edu, aczajka@nd.edu}
}
\maketitle

\input{latex/sections/00-abstract-v1}
\input{latex/sections/01-introduction-v1}
\input{latex/sections/02-related-work-v1}
\input{latex/sections/03-methodology-v3}
\input{latex/sections/04-results-v3}
\input{latex/sections/06-conclusion-v1}

\section*{Acknowledgments}

This work was supported by the U.S. Department of Defense (Contract No. W52P1J-20-9-3009) and by the National Science Foundation (Grant No. 2237880). Any opinions, findings, and conclusions or recommendations expressed in this material are those of the authors and do not necessarily reflect the views of the National Science Foundation, the U.S. Department of Defense or the U.S. Government. The U.S. Government is authorized to reproduce and distribute reprints for Government purposes, notwithstanding any copyright notation here on. We would also like to acknowledge and extend our thanks to Jacob Piland for his work in curating and preparing the chest X-ray dataset featured in this paper.
\small
\bibliographystyle{ieee_fullname}
\bibliography{egbib}

\end{document}

%% file: latex/sections/00-abstract-v1.tex
\begin{abstract}
Leveraging human perception into training of convolutional neural networks (CNN) has boosted generalization capabilities of such models in open-set recognition tasks. One of the active research questions is where (in the model architecture or training pipeline) and how to efficiently incorporate always-limited human perceptual data into training strategies of models. In this paper, we introduce MENTOR (huMan pErceptioN-guided preTraining fOr increased geneRalization), which addresses this question through two unique rounds of training CNNs tasked with open-set anomaly detection. First, we train an autoencoder to learn human saliency maps given an input image, without any class labels. The autoencoder is thus tasked with discovering domain-specific salient features which mimic human perception. Second, we remove the decoder part, add a classification layer on top of the encoder, and train this new model conventionally, now using class labels. We show that MENTOR successfully raises the generalization performance across three different CNN backbones in a variety of anomaly detection tasks (demonstrated for detection of unknown iris presentation attacks, synthetically-generated faces, and anomalies in chest X-ray images) compared to traditional pretraining methods (e.g., sourcing the weights from ImageNet), and as well as state-of-the-art methods that incorporate human perception guidance into training. In addition, we demonstrate that MENTOR can be flexibly applied to existing human perception-guided methods and subsequently increasing their generalization with no architectural modifications.
\end{abstract}

%% file: latex/sections/01-introduction-v1.tex
\section{Introduction}
\label{sec:introduction}
\begin{figure}[!htb]
  \centering
  \includegraphics[width=\linewidth]{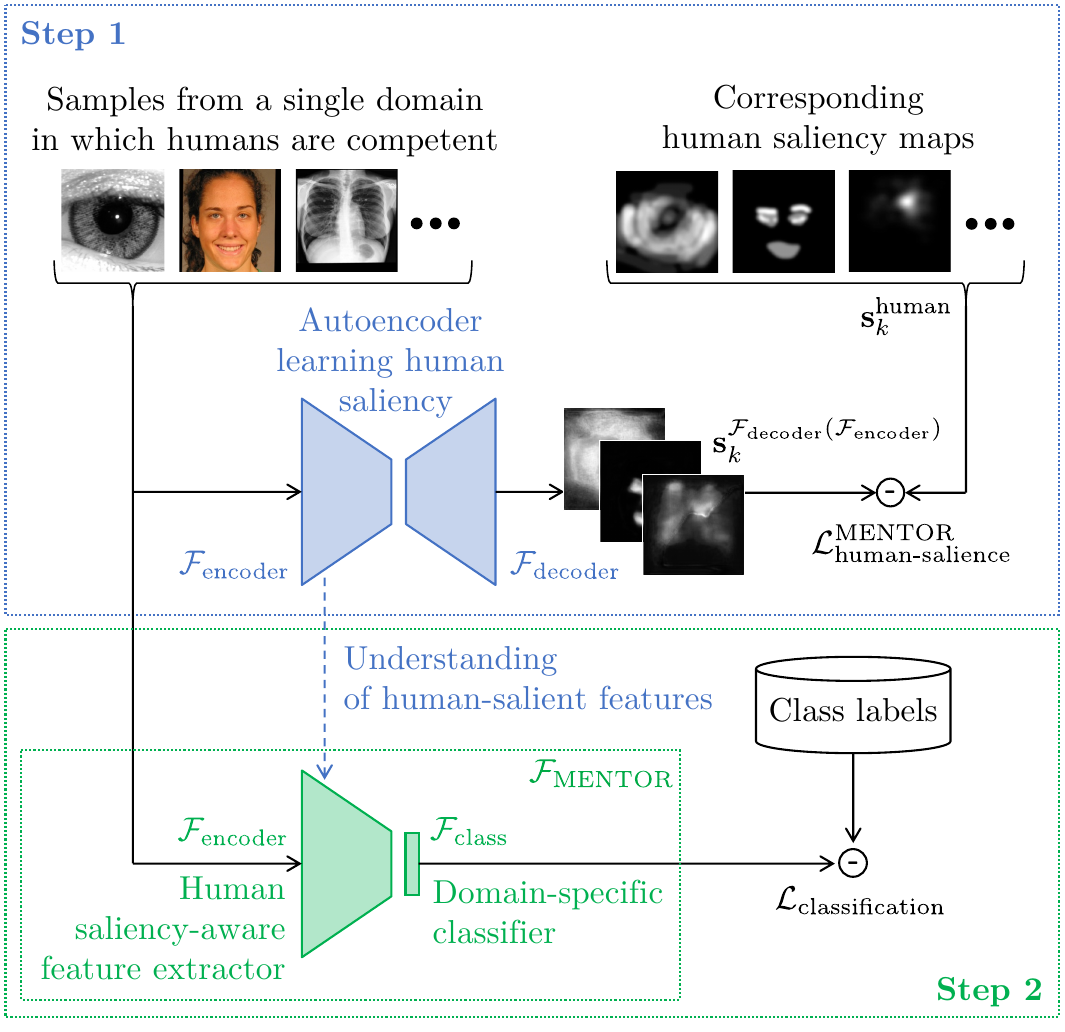}
  \caption{MENTOR approach. \textcolor{mentor-blue}{Step 1: An autoencoder-based model is first trained to recreate human saliency maps, and thus to build an understanding of human perception-sourced salient features into the encoder $\mathcal{F}_\text{encoder}$.} \textcolor{mentor-green}{Step 2: Such pre-trained encoder $\mathcal{F}_\text{encoder}$ is then decoupled from the autoencoder, and along with a classifier $\mathcal{F}_\text{class}$ are tuned for the anomaly detection task utilizing standard cross-entropy loss.} MENTOR has been designed and evaluation to address generalization required in a {\bf single domain} (in this paper independently for iris presentation attack, synthetic face and chest X-ray-based anomaly detection).}
  \label{fig:teaser}
\end{figure}

Human perceptual information is often incorporated into deep learning training strategies in order to improve generalization \cite{boyd2022human, parzianello2022saliency}, align models with human-sourced saliency \cite{boyd2021cyborg, czajka2019domain}, and reduce training time by supplying additional prior knowledge into the training process \cite{boyd2021cyborg, van2023probabilistic}. However, a common and valid criticism is the high cost of acquiring human annotations or eye tracking-sourced saliency data. For anomaly detection tasks, such as biometric presentation attack detection (PAD), new attack types (\ie paper print outs, doll eyes, textured contact lens) are developed frequently.
Medical diagnostics may require continual update of human perceptual understanding of anomalies present in medical samples. In all these tasks, requiring few-shot learning solutions functioning well in an open-set classification regime, collecting more training samples and/or more human saliency data is difficult or even impossible. This is why an effective use of the existing, although always-limited human saliency information originating from domain experts is crucial. An associated question is \emph{how} and \emph{where} to appropriately incorporate human perceptual understanding of a task into model training to deliver strong priors allowing for better generalization of such models.

This paper introduces MENTOR, a framework for human perception-guided pretraining, evaluated in the context of open-set iris PAD, synthetic face detection, and disease classification from chest X-ray scans. 
MENTOR relies on the intuition that (as children) we first learn the representations of the visual world without knowing the ``class labels,'' and we assign meaning in what we see later in our lives. Around being one year old, infants began to point, imitate, and understand objects merely by observation or visual salience, without any explicit knowledge of the object's application or type (classification) \cite{gopnik1999scientist}. We emulate this iterative process by first training an autoencoder to learn the associations between an input image and a human saliency map (only visual salience, with no class labels). Once this representation is learned, the decoder-side is removed, a classifier is put on top of the encoder's latent space, and the entire model is then fine-tuned in a regular way using cross-entropy loss to solve a classification task at hand. Fig. \ref{fig:teaser} illustrates this pipeline. Our experiments show that such class label-agnostic, but human perception-aware pretraining of the model's backbone allows for (a) better generalization to unknown samples than models fine-tuned in the same way but initialized with weights obtained in pretraining on a massive amount data such as ImageNet, (b) same as (a) when compared to models trained with a state-of-the-art human perception guidance incorporated into a loss function. We make these observations for three different domains, in which humans can deliver meaningful saliency information (iris PAD, synthetic face detection, and chest X-ray-based diagnosis), and for three different neural network architectures (ResNet, Inception and EfficientNet). In other words, MENTOR efficiently incorporates the limited human perceptual understanding of a domain into training of convolutional neural networks (CNN).

Two core differences between MENTOR and state-of-the-art unsupervised representation learning approaches, such as DINO \cite{Caron_ICCV_2021}, are that (a) MENTOR leverages human perception-based guidance in the pre-training phase, and (b) MENTOR operates in a very data-limited regime, since it's targeted to domains, in which collecting data is intricate, but humans competent in a given domain can be found (\ie physicians in medical diagnosis). It is also noteworthy that MENTOR works with any CNN architecture, and without any new or invasive architectural changes. This paper is organized around the following two {\bf research questions}:
\begin{itemize}
    \item \textbf{RQ1:} Does MENTOR improve the performance of iris presentation attack detection, synthetic face detection, and chest X-ray-based diagnosis compared to conventional vision pretraining techniques \cite{ImageNet} and human perception-guided methods? (answered in \emph{Sec. \ref{sec:RQ1})}
    \item \textbf{RQ2:} Can MENTOR be applied to existing human perception-guided techniques to further increase their generalization? (answered in \emph{Sec. \ref{sec:RQ2}})
\end{itemize}

The sources codes of the proposed approach are offered at \url{https://github.com/CVRL/MENTOR}

%% file: latex/sections/02-related-work-v1.tex
\section{Related Work}
\label{sec:related-work}

\subsection{Human Saliency-Guided Training} Human perceptual information related to visual tasks is usually collected via image annotations \cite{boyd2022human, boyd2021cyborg}, eye tracking \cite{boyd2023patch, czajka2019domain}, or measuring reaction times \cite{huang2023measuring}. Human salience, estimated from the perceptual data, has been successfully integrated into the training process by perception-based training data augmentations \cite{boyd2022human}, integrating the perceptual information into the loss function \cite{boyd2021cyborg, huang2023measuring, piland2023model}, or as a regularization approach \cite{dulay2022using}. Other approaches include using attention mechanisms, which typically require changes to the model and are architecture-specific \cite{linsley2018learning}. Human-guided models have shown to improve performance \cite{boyd2021cyborg}, model interpretability \cite{linsley2018learning} and explainability \cite{crum2023explain}.

The most architecture-agnostic implementations of incorporating human saliency into model's training are those using perception-specific loss function components. Boyd \etal \cite{boyd2021cyborg} introduced the CYBORG loss function, which penalizes the model for divergence between human saliency maps and model's Class Activation Maps (CAM) \cite{zhou2016learning}. While this method aligns the activations of the model's last convolutional layer with human saliency, it's unclear if that mechanism actually builds a human perceptual intelligence into the model, especially in its earlier layers, or acts as a regularizer by lowering the entropy of model's salience \cite{piland2023model}. Other methods use autoencoding strategies, such as UNET+Gaze to incorporate human salience into training a classifier, though these operate as somewhat independent tasks \cite{karargyris2021creation}. These methods train the classification and human-salient loss components jointly, which may often conflict with one another during training, leaving to information collapse during the reconstruction phase, or forcing the model to obey only one loss term at a time. Instead, the MENTOR approach proposed in this paper is different from UNET+Gaze and CYBORG (as well as other similar human perception-based training strategies) in a sense that it encourages the model to first build associations between input images and human-sourced salient features, without telling the model what task these features are useful for, to create more general interpretations of the visual world. This allows the two tasks to be fully disentangled so that generalization capabilities can be built upon existing feature representations learned from the previous training step. 

Sonsbeek \etal utilize a two-stage training strategy to learn global and local features \cite{van2023probabilistic}. Their model is first trained on a larger dataset with no human salience, then fine-tuned on a smaller dataset with human saliency maps using a knowledge distillation module, with freezing selected weights. MENTOR deploys the reverse strategy, leveraging human saliency maps first for learning class-agnostic human-sourced features, and in its second stage utilizing a regular cross-entropy-based training without a requirement to freeze any weights.

\subsection{Efficient Use of Human Annotations} Crum \textit{et al.} \cite{crum2023teaching} applied human salience in a Teacher-Student training paradigm to make a better use of limited human salience data. In their work, human annotations were first used to train ``teacher'' models, which are then used to generate subsequent model salience for training ``student'' models. Teachers were trained using the CYBORG loss, and afterwards used to generate saliency maps on un-annotated training samples using CAM \cite{zhou2016learning} and RISE \cite{petsiuk2018rise}. This training paradigm boosted performance in synthetic face detection and iris PAD. Interestingly, the MENTOR's byproduct is an autoencoder that generates human saliency maps for unseen data, and thus it can complement approaches such as the above teacher-student learning paradigms. However, challenges with the Teacher-Student paradigm involve which is the optimal ``teacher'' model to select, and obtaining sufficient human-salient training samples for the Teachers to learn the task and subsequently train Student models.

%% file: latex/sections/03-methodology-v3.tex
\section{MENTOR Approach}
\label{sec:methodology}
\subsection{Methodology} 

\input{latex/figures/autoencoder-all-datasets}
MENTOR is a novel approach of incorporating human salience into model training through a two-part training series. First, an autoencoder is trained to generate human-like saliency given an input image, but without any class labels (``Step 1'' in Fig. \ref{fig:teaser}). In that way, the model creates associations between input samples and human-salient regions.
Once these associations are made, we put a single, fully-connected layer (initialized with random weights) on top of the encoder's embeddings and fine-tune the entire architecture to solve the classification task (see ``Step 2'' in Fig. \ref{fig:teaser}).

It's analogous to theories involving how humans gain perceptual understanding of the visual world, by first being exposed to  stimuli without being given explanations (or ``class labels'') of what they see. As shown in Sec. \ref{sec:results}), such human perception-based pre-training results in a better performance in three anomaly detection tasks (from different domains) compared to those observed for models initialized with weights obtained after training with a large visual dataset (ImageNet), and than for models trained with state-of-the-art loss function-based human perception guidance.

More formally, let's consider a state-of-the-art family of approaches, such as \cite{boyd2021cyborg,Ismail_NeurIPS_2021,Pham_WACV_2024}, incorporating human salience into loss functions used to train a model $\mathcal{F}$ in a supervised manner in one phase. Re-phrasing the CYBORG loss \cite{boyd2021cyborg} (helpful in stressing the differences between the above approaches and MENTOR), we can summarize human-guided loss functions as:

\begin{equation}
\mathcal{L} = \alpha\mathcal{L}_\text{classification} + (1-\alpha)\mathcal{L}_\text{human-salience}
\end{equation}

\noindent
where $\alpha$ is a trade-off parameter weighting human- and model-based saliencies (not needed in MENTOR, as seen later), while

\begin{equation}
\mathcal{L}_\text{classification} = \frac{1}{K}\sum_{k=1}^K\sum_{c=1}^{C}\bm{1}_{y_k \in \mathcal{C}_c}\Big(\log p_{\mathcal{F}}(y_k \in \mathcal{C}_c)\Big)
\label{eq:cyborg_classification}
\end{equation}

\noindent
is a cross-entropy-based classification loss function, and

\begin{equation}
\mathcal{L}_\text{human-salience} = \frac{1}{K}\sum_{k=1}^K\sum_{c=1}^{C}\mathcal{D}\Big(\textbf{s}_{k,\mathcal{C}_c}^{\text{human}} - \textbf{s}_{k,\mathcal{C}_c}^{\mathcal{F}}\Big)
\label{eq:cyborg_human}
\end{equation}

\noindent
where $y_k$ is a class label for the $k$-th sample, $\bm{1}$ is an indicator function equal to $1$ when $y_k \in \mathcal{C}_c$, and 0 otherwise, $C$ is the total number of classes, $K$ is the number of samples in a batch, $\textbf{s}_{k,\mathcal{C}_c}^{\text{human}}$ and $\textbf{s}_{k,\mathcal{C}_c}^{\mathcal{F}}$ are the human and the model $\mathcal{F}$'s saliency for the $k$-th sample representing class $\mathcal{C}_c$, respectively, and finally $\mathcal{D}$ is the measure of dissimilarity between saliency maps (\eg the Mean Square Error distance). 

MENTOR, in contrast to the above formulation, disentangles the human-perception-guided pre-training and classification fine-tuning of the model $\mathcal{F}$ consisting of two components: the encoder part $\mathcal{F}_\text{encoder}$, learning human salience, and classification part $\mathcal{F}_\text{class}$. Namely, in the \textcolor{mentor-blue}{\bf first step} (cf. Fig. \ref{fig:teaser}) we use the human saliency to train $\mathcal{F}_\text{encoder}$ using the following $\mathcal{L}_\text{human-salience}^\text{MENTOR}$ loss:

\begin{equation}
\mathcal{L}_\text{human-salience}^\text{MENTOR} = \frac{1}{K}\sum_{k=1}^K\big\|\textbf{s}_k^{\text{human}} - \textbf{s}_k^{\mathcal{F}_\text{decoder}(\mathcal{F}_\text{encoder})}\big\|^2
\end{equation}

\noindent
where $\|\cdot\|$ is the $\ell_2$ norm, and $\textbf{s}_k^{\mathcal{F}_\text{decoder}(\mathcal{F}_\text{encoder})}$ is the saliency map generated by the model for the $k$-th sample. Note that no class labels are used in pre-training. In the \textcolor{mentor-green}{\bf second step} (cf. again Fig. \ref{fig:teaser}), only the cross-entropy loss \eqref{eq:cyborg_classification} is used, in which $\mathcal{F}(\cdot)=\mathcal{F}_\text{MENTOR}(\cdot) = \mathcal{F}_\text{class}\big(\mathcal{F}_\text{encoder}(\cdot)\big)$. We start with random weights in $\mathcal{F}_\text{class}$, and do not freeze the weights of $\mathcal{F}_\text{encoder}$ in Step 2, hence the entire model is fine-tuned. 

Note that the decoder part $\mathcal{F}_\text{decoder}$ is not utilized in Step 2, however the entire autoencoder trained in Step 1 can also be used to generate human-like salience maps and replace real salience maps in previosly-proposed human saliency-based training paradigms, such as \cite{boyd2021cyborg,crum2023teaching,Ismail_NeurIPS_2021} or \cite{Pham_WACV_2024}.

\subsection{Neural Network Architectural Choices}
The MENTOR approach does not specify a type of autoencoder. We experimented with UNET \cite{ronneberger2015u} and UNET++ \cite{zhou2018unet++} with three different backbones: ResNet152 \cite{he2016deep}, Inception-V4 \cite{szegedy2017inception} and EfficientNet-b7 \cite{tan2019efficientnet}) using the same training configurations \cite{Iakubovskii:2019} to observe stability of the proposed approach across visual domains and architectures.

Specific domains and backbones for the autoencoder reported in Sec. \ref{sec:results} are as follows: UNET (all backbones for iris PAD domain) and UNET++ (all backbones for synthetic face detection and chest X-ray-based diagnosis).

Also, MENTOR does not specify the classifier put in Step 2 on top of the encoder embeddings. We demonstrate effectiveness of this approach by adding a single-layer classifier to minimize the extra architectural components added to human saliency-guided pre-trained encoder, but there are no theoretical reasons for not exploring the generalization gain achieved for deeper structures as a future research.

\subsection{Training and Evaluation Details}
MENTOR does not need any specific hyperparameter settings. For reference purposes, we provide settings we have used in this work. The autoencoders in Step 1 were trained with the AdamW optimizer, with a learning rate of 0.0001, and with a batch size of 8. MSE loss was used to assess the agreement bewteen  human saliency and the predicted saliency maps, all scaled to a canonical $224\times224$ pixel resolution. Training was continued until 50 epochs, but many models converged quickly (in less than 5 epochs). All full models in Step 2 were trained using cross-entropy loss using Stochastic Gradient Descent (SGD), with a learning rate of 0.005 decreased by 0.1 every 12 epochs. As in Step 1, maximum number of epochs was 50. CYBORG and UNET+Gaze models were trained from the source codes and training configurations provided by their work \cite{boyd2021cyborg, karargyris2021creation}.

We evaluate the proposed MENTOR approach using Area Under the Receiver Operating Characteristic Curve (AUROC). For all experiments we perform ten independent training runs instantiated using different seeds.

\section{Datasets}
\label{sec:datasets}
\subsection{Training and Validation Data Subsets}

\paragraph{Domain 1: Iris Presentation Attack Detection} For training the autoencoder (``Step 1'' in Fig. \ref{fig:teaser}), we use the only (known to us) dataset of {\it bona fide} and anomalous iris images accompanied by human saliency maps, collected by Boyd \etal \cite{boyd2022human}. The {\it bona fide} and abnormal iris images for this experiment were sampled from a superset of already published live iris and iris presentation attack sets \cite{casia-database,Sung_OE_2007,Galbally_ICB_2012,Kohli_ICB_2013,Yambay_ISBA_2017,Trokielewicz_IVC_2020,Kohli_BTAS_2016,Wei_ICPR_2008,Trokielewicz_BTAS_2015,Yambay_IJCB_2017,Das_IJCB_2020}. In Boyd \etal's experiments, participants recruited via Mechanical Turk were asked whether the iris was {\it bona fide} or abnormal, and were instructed to hand-annotate regions of the image which support their decision. Only correctly classified samples were used in post-processing. Annotations of the same image originating from multiple subjects were averaged together, resulting in 765 saliency maps (see Fig. \ref{fig:annotator} for examples).
These 765 images and saliency maps were randomly divided by the authors into train (n=612) and validation (n=153) sets, and this split was used in training of all the autoencoder instances in this work.

For training the subsequent iris PAD classification models (``Step 2'' in Fig. \ref{fig:teaser}), the same 765 images (but  with class labels and no salience maps) were used as training images, with additional 23,312 validation images sampled from the same superset as used in \cite{boyd2022human}. The train and validation sets are subject-disjoint. The 765 training samples were comprised of 198 live irises and 567 spoof irises.

\paragraph{Domain 2: Synthetic Face Detection} We trained the autoencoder using real and synthetically-generated face images and human saliency maps offered by \cite{boyd2021cyborg}. 363 humans recruited via Amazon Mechanical Turk annotated regions of the image deemed important to their classification decision (real / fake). Only correctly classified samples (images and accompanying saliency maps) were used during training. For simplicity, 765 samples (379 real / 386 synthetic) were randomly selected without replacement to match the number of samples available for Domain 1 (Iris PAD).

For training the subsequent synthetic face classification models (``Step 2'' in Fig. \ref{fig:teaser}), all 765 images were used as training images in conjunction with additional 20,000 validation samples extracted from the same sources as in \cite{boyd2021cyborg}. Train and validation sets were subject-disjoint.

\paragraph{Domain 3: Chest X-ray-based Diagnosis} The autoencoder was trained with normal and abnormal chest X-ray images \cite{johnson2019mimic}, and corresponding eye-tracking visual salience collected from radiologists examining these X-ray scans \cite{bigolin2022reflacx}. We consider only abnormal categories represented by at least 200 training images, that is: atelectasis, cardiomegaly, edema, lung opacity, pleural effusion, pneumonia, and support devices. We removed images containing erroneous class labels and consider only images where radiologists were certain of their classification decision. The intensity of pixels in eye tracking-based salience maps is proportional to the strength of the eye fixation. If a sample included human saliency maps from multiple radiologists, we combined these maps by taking the maximum intensity for each pixel. This was done to preserve salient information provided by each expert, who may consider different features when classifying the samples, instead of considering only the most ``popular'' features. All images were resized to $224\times224$. For consistency, 765 samples (223 normal / 542 abnormal) and accompanying saliency maps were used during the training phases to match with sample sizes used in two other domains.

For training the chest X-ray classification models (``Step 2'' in Fig. \ref{fig:teaser}), all 765 images were used as training images, in conjunction with additional 4,878 subject-disjoint validation samples sampled from the same dataset.

\subsection{Test Data Sets}

\paragraph{Domain 1: Iris Presentation Attack Detection} We evaluate MENTOR using the Iris Liveness Detection 2020 Competition (LivDet-2020) official test set \cite{das2020iris}. LivDet-2020 is a competition held to benchmark iris presentation attack algorithms, and 2020 edition included the largest number of abnormal (from iris recognition point of view) classes, including: artificial eyeballs with irises printed on them, textured contact lenses, postmortem iris images, paper print outs, synthetically-generated iris samples, images of diseased eyes, and images of eyes wearing textured contact lenses printed on paper and then re-photographed. The evaluation made with the LivDet-Iris 2020 test data is thus quite rigorous as it tests the algorithms in an open-set scenario, in which attack types unseen during training are used in testing.

\paragraph{Domain 2: Synthetic Face Detection} We evaluate MENTOR models detecting synthetic faces using (a) synthesized images sampled from six different GAN architectures (ProGAN \cite{karras2017progressive}, StarGANv2 \cite{stargan}, StyleGAN \cite{styleGAN}, StyleGAN2 \cite{styleGAN2}, StyleGAN2-ADA \cite{syleGAN2-ADA}, and StyleGAN3 \cite{styleGAN3}, and (b) authentic face images from FFHQ \cite{FFHQ} and CelebA-HQ \cite{Celeb}, comprising of 7,000 test images. These GAN models and the source of live faces were not considered in generating the train and validation sets, hence the open-set regime of this evaluation protocol is kept. Future works may include testing with additional synthetic face image benchmarks, incorporating for instance stable diffusion-based samples \cite{Rombach_CVPR_2022}, if such formal benchmarks are available.

\paragraph{Domain 3: Chest X-ray-based Diagnosis} The test set was comprised of 8,265 samples extracted from \cite{johnson2019mimic}, completely disjoint from previous training splits. Samples with at least one abnormal category (atelectasis, cardiomegaly, edema, lung opacity, pleural effusion, pneumonia, or support devices) were classified as abnormal, with the contrary being normal samples.

%% file: latex/figures/autoencoder-all-datasets.tex
\begin{figure}[t]
  \centering
  \includegraphics[width=0.75\linewidth]{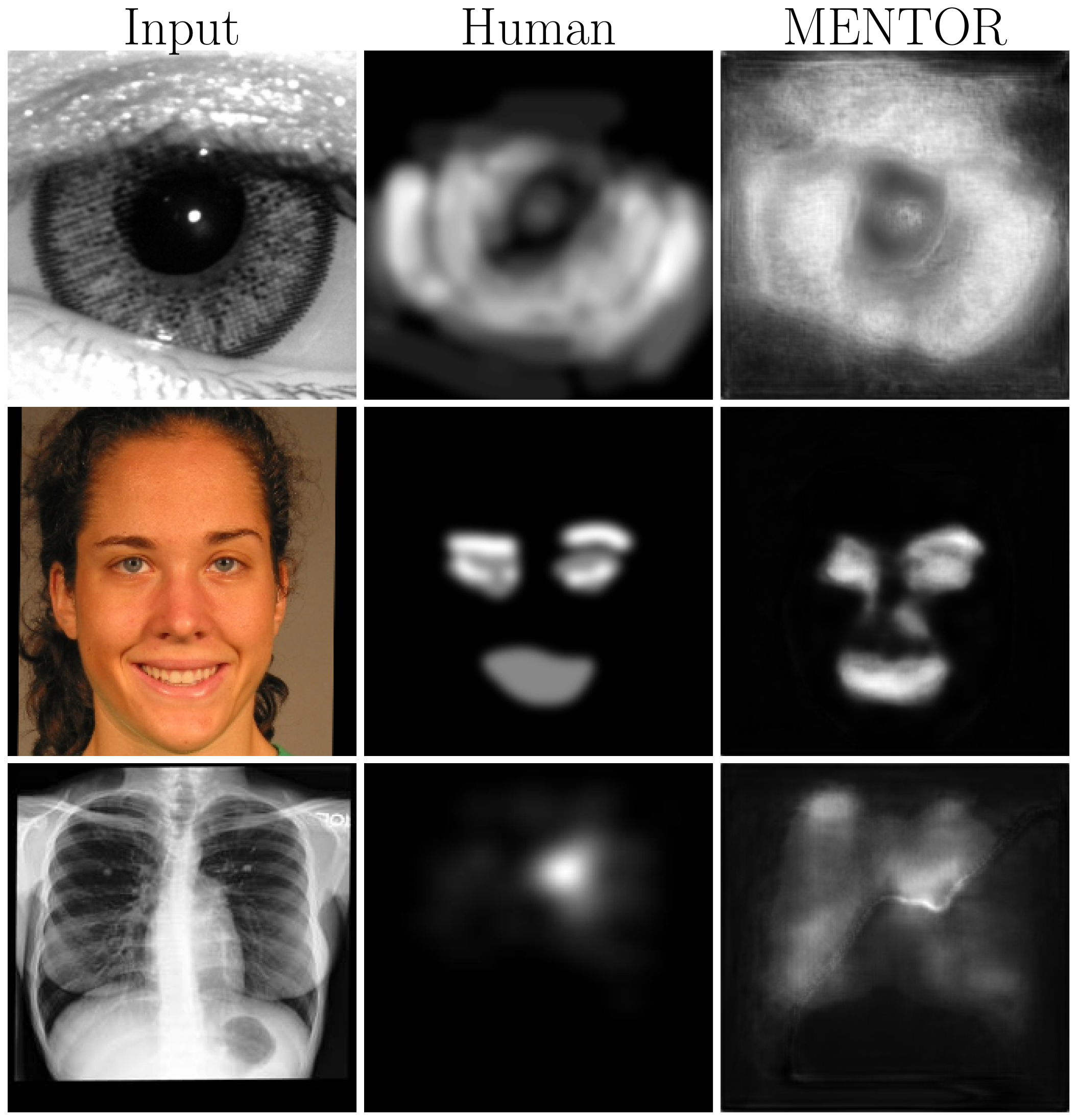}
  \caption{A byproduct of the MENTOR pre-training approach is the autoencoder $\mathcal{F}_\text{decoder}\big(\mathcal{F}_\text{encoder}(\cdot)\big)$ predicting saliency maps that resemble human salience (\textbf{top row:} iris presentation attack detection, \textbf{middle row:} synthetic face detection, and \textbf{bottom row:} chest X-ray-based diagnosis).}
  \label{fig:annotator}
\end{figure}

%% file: latex/sections/04-results-v3.tex
\section{Results}
\label{sec:results}
\input{latex/tables/models-methods-v4}
\input{latex/tables/mentor-plus-methods}
To answer both research questions, we compare MENTOR-trained models with those (a) pretrained using a classical learning approach, initialized with ImageNet weights, and (b) trained with a state-of-the-art human perception-guided approaches using the CYBORG loss \cite{boyd2021cyborg} and UNET+Gaze \cite{karargyris2021creation}.

\subsection{Improvement of Generalization}
\label{sec:RQ1}

MENTOR pretraining compares very favorably to models initialized with ImageNet weights, as well as to models trained with a state-of-the-art human-perception training paradigms using the same data and the same human saliency maps (see Tab. \ref{tab:performance}). MENTOR achieves improved average performance in all but one training configuration in Synthetic Face Detection (ResNet152 trained using cross-entropy, AUC=0.540), where it was marginally worse (0.539). MENTOR displayed high performance for Iris PAD across all backbones, with the best being ResNet. Inception and EfficientNet were equally strong in raising generalization performance with MENTOR (0.899 and 0.901, respectively), with CYBORG (0.794) and UNET+Gaze (0.807) performing the poor on Inception-based backbones, even lagging behind cross-entropy (0.879). MENTOR performed well in synthetic face detection, offering large improvements over existing saliency-based methods (CYBORG, UNET+Gaze), who struggled to attain above random chance. These findings suggest that existing human saliency training methods used in synthetic face detection struggle in smaller training sets (only 765 samples). Finally, MENTOR showed consistently strong performance in chest x-ray diagnosis across all models using MENTOR, with the largest gains compared to existing methods coming from Inception. \textbf{Thus, answering RQ1, we conclude that MENTOR pretraining is an effective method to improve the generalization capabilities in iris PAD, synthetic face detection, and chest x-ray-based diagnosis tasks.}

\subsection{Combining MENTOR with other methods}
\label{sec:RQ2}

To answer RQ2, we evaluated two human-guided training methods (CYBORG \& UNET+Gaze) using MENTOR's pretrained weights.
As seen in Tab. \ref{tab:mentor-plus-method}, MENTOR greatly increases the generalization performance of both methods across all domains. MENTOR pretraining coupled with the CYBORG loss function resulted in a 0.043 increase in Iris PAD, and 0.041 increase for synthetic face detection. UNET+Gaze had a substantial benefit in the chest x-ray diagnosis task. These results indicate that MENTOR pretraining provides human perception-guided methods with salient-aware kernels that more easily adapt to a better solution space than conventional methods. In addition, these results indicate MENTOR's strengths in complementing existing saliency-based training strategies, further contributing to its novelty and expanding its potential widespread use. \textbf{Thus, answering RQ2 we conclude that MENTOR pretraining can be easily adapted to other saliency-based methods to increase the generalization performance in the three domains considered in this work.}

\input{latex/tables/smaller-models-v3}

\section{Ablation Study}
In this section, we provide extended experimentation and analysis of MENTOR under a variety of different benchmarks and scenarios. While these experiments are not core to our research contributions, they bolster the overall capabilities of our method. First, we benchmark MENTOR against the Teacher-Student Paradigm, which is a SOTA method used to extend the limited uses of human saliency. Second, we investigate both qualitatively and quantitatively the saliency maps offered by each of the methods. Finally, we explore MENTOR's capabilities on a variety of smaller ResNet backbones.

\subsection{MENTOR \& Teacher-Student Framework}
\input{latex/tables/teacher-student-v3}
Although our focus was on raising the generalization performance in limited training samples, MENTOR's two round training strategy can be utilized to synthetically generate saliency for subsequent training samples. To explore this idea, we compared MENTOR's salience generated from Step 1 the Teacher-Student paradigm \cite{crum2023teaching}. Due to time and compute constraints, we included experiments for the Iris PAD task following the same configurations and training splits offered by \cite{crum2023teaching} (see Tab. \ref{tab:teacher-student}). Without \emph{any} fine-tuning, ``student'' models trained using MENTOR's saliency was able to outperform SOTA performance in half of the published architectures, with dominant gains with ResNet50 (0.920 to 0.941) and marginal improvement with Inception-V3 (0.947 to 0.947). MENTOR remained competitive for Xception-based architectures (0.952 to 0.947), but lagged behind for DenseNet (0.950 to 0.931). Though not a main focus of our work, we are able to indicate the versatility of our method.

\subsection{Analysis of Saliency Maps}
\input{latex/figures/saliency-sampler}

Finally, we included both a qualitative and quantitative analysis comparing the saliency generated by MENTOR and the other methods included in this paper.

\paragraph{Qualitative evaluation} An extended visualization of Fig. \ref{fig:annotator} can be found in Fig. \ref{fig:extended-annotator} for more examples. As discussed previously (Sec. \ref{sec:methodology}), saliency-based training is traditionally trained jointly, where the network is tasked with satisfying both the classification and human-salient (\ie often reconstruction, or minimizing the difference between human and model saliency maps) components. However, with more strenuous tasks, the network is often forced to make a trade off between classification performance or the adherence towards human saliency (otherwise simply holistic task understanding). Occasionally, these components may altogether prove to be incompatible with existing techniques when trained jointly. Fig. \ref{fig:extended-annotator} shows UNET+Gaze suffering from information collapse during its reconstruction of the human saliency map, and instead favoring higher classification performance. Likewise, CYBORG produces a low-resolution saliency map that misses key features found within the human annotation. Instead, MENTOR alleviates these concerns by employing a two-round training strategy, where first the task-specific representations are built into the model (resulting in high reconstruction quality), and second associations are made between these feature representations and the classification labels. 

\input{latex/tables/ablation-entropy}
\paragraph{Quantitative evaluation}In addition, we included some quantitative statistics involving the saliency maps generated from cross-entropy, CYBORG, and MENTOR trained models offered by \cite{crum2023explain}. These measures provide a toolkit to access the explainability of model salience, especially within the context of synthetic face detection. Due to time constraints, we measured the Salience Entropy ($S_{entropy}$) of the Class Activation Mappings (CAM) generated on the test set \cite{crum2023explain, zhou2016learning}. Salience Entropy attempts to measure the compactness or focus of the saliency maps. Initial observations appear to indicate that MENTOR may have slightly higher average $S_{entropy}$ than existing saliency-based methods, contrary to previous findings in synthetic face detection \cite{crum2023explain}. However, due to the sparsity of saliency (\ie eye-tracking associated with chest x-rays, and attack-type specific annotations for iris PAD related tasks, new ways of quantitatively comparing saliency may be necessary. In this work, we used all available human saliency during training, so there is simply no way to measure the efficacy between human and model salience without collecting additional data, saving this for future work.

\subsection{MENTOR on Smaller Architectures} Our core findings suggest that MENTOR can be an effective strategy to raise generalization in larger CNN models. We also included additional experimentation that support MENTOR's ability to raise generalization in smaller models (see Tab. \ref{tab:resnet-ablation-study}). MENTOR provides the highest gains in generalization performance across ResNet152 and ResNet50 backbones, while being competitive with the CYBORG loss for ResNet32 and ResNet18 for Iris PAD and Chest x-ray anomalies. Note that no additional changes in training configurations were made between all ResNet experiments. For future experiments, likely changes with the learning rate will be the most beneficial for using MENTOR on smaller models (otherwise using different optimizers). As shown previously with RQ2, MENTOR pairs well with strong regularization loss components (see Tab. \ref{tab:mentor-plus-method}). Likely a combination between MENTOR and loss function components (\ie CYBORG) could greatly raise the generalization performance depending on the task.

%% file: latex/tables/models-methods-v4.tex
\begin{table*}[!htb]
\centering
\scriptsize
\caption{Mean and standard deviations of AUROC scores (calculated independently for each combination of domain-architecture) across 10 independent runs. The \fbox{\textcolor{red}{\textbf{\xmark}}} indicates that no human saliency was used during training, whereas \fbox{\textcolor{ForestGreen}{\textbf{\cmark}}} indicates human saliency was used during training.}
\label{tab:performance}
\begin{tabular}{@{}l|>{\columncolor[gray]{0.95}}c>{\columncolor[gray]{0.95}}c>{\columncolor[gray]{0.95}}l|l@{}}
\toprule
\textbf{Dataset} & \textbf{Cross-entropy} & \textbf{CYBORG \cite{boyd2021cyborg}} & \textbf{UNET+Gaze \cite{karargyris2021creation}} & \textbf{MENTOR (ours)} \\
\text{\hspace{5mm}Trained Using Human Saliency} & \textcolor{red}{\textbf{\xmark}} & \textcolor{ForestGreen}{\textbf{\cmark}} & \hspace{7mm}\textcolor{ForestGreen}{\textbf{\cmark}} & \hspace{7mm}\textcolor{ForestGreen}{\textbf{\cmark}} \\
\midrule
\textbf{Iris PAD} & & & \\
\hspace{5mm}ResNet152 & 0.879$\pm$0.02 & 0.897$\pm$0.02 & 0.892$\pm$0.03 & \textbf{0.912$\pm$0.02} \\
\hspace{5mm}Inception-V4 & 0.879$\pm$0.03 & 0.794$\pm$0.04 & 0.807$\pm$0.024 & \textbf{0.899$\pm$0.05} \\
\hspace{5mm}EfficientNet-7b & 0.881$\pm$0.01 & 0.801$\pm$0.03 & 0.830$\pm$0.039 & \textbf{0.901$\pm$0.02} \\
\textbf{Synthetic Face Detection} & & & & \\
\hspace{5mm}ResNet152 & \textbf{0.540$\pm$0.04} & 0.508$\pm$0.05 & 0.432$\pm$0.042 & \text{0.538$\pm$0.05} \\
\hspace{5mm}Inception-V4 & 0.519$\pm$0.06 & 0.484$\pm$0.03 & 0.438$\pm$0.053 & \textbf{0.585$\pm$0.03} \\
\hspace{5mm}EfficientNet-7b & 0.465$\pm$0.02 & 0.500$\pm$0.03 & 0.453$\pm$0.086 & \textbf{0.565$\pm$0.03} \\
\textbf{Chest X-ray-based diagnosis} & & & \\
\hspace{5mm}ResNet152 & 0.840$\pm$0.01 & 0.854$\pm$0.01 & 0.840$\pm$0.011 & \textbf{0.856$\pm$0.01} \\
\hspace{5mm}Inception-V4 & 0.840$\pm$0.01 & 0.830$\pm$0.02 & 0.791$\pm$0.018 & \textbf{0.851$\pm$0.00} \\
\hspace{5mm}EfficientNet-7b & 0.817$\pm$0.01 & 0.820$\pm$0.01 & 0.829$\pm$0.009 & \textbf{0.838$\pm$0.01} \\
\bottomrule
\end{tabular}
\end{table*}

%% file: latex/tables/mentor-plus-methods.tex
\begin{table}[!htb]
\centering
\scriptsize
\caption{MENTOR pretraining combined with each baseline method. Means and standard deviations of AUROC scores are reported across 10 independent runs using a ResNet152 backbone.}
\label{tab:mentor-plus-method}
\begin{tabular}{@{}l|>{\columncolor[gray]{0.95}}c|l@{}}
\toprule
\textbf{Initialization Weights} & \textbf{ImageNet \cite{ImageNet}} & \textbf{MENTOR (ours)} \\
\midrule
\textbf{Iris PAD} & \textbf{} & \textbf{} \\
\hspace{3mm}CYBORG \cite{boyd2021cyborg} & \text{0.897$\pm$0.020} & \textbf{0.940$\pm$0.026} \\
\hspace{3mm}UNET+Gaze \cite{karargyris2021creation} & \text{0.892$\pm$0.030} & \textbf{0.914$\pm$0.017} \\
\textbf{Synthetic Face Detection} & \textbf{} & \textbf{} \\
\hspace{3mm}CYBORG \cite{boyd2021cyborg} & 0.508$\pm$0.050 & \textbf{0.549$\pm$0.014} \\
\hspace{3mm}UNET+Gaze \cite{karargyris2021creation} & 0.432$\pm$0.042 & \textbf{0.439$\pm$0.059} \\
\textbf{Chest X-ray-based diagnosis} & \textbf{} & \textbf{} \\
\hspace{3mm}CYBORG \cite{boyd2021cyborg} & \text{0.854$\pm$0.010} & \textbf{0.867$\pm$0.006} \\
\hspace{3mm}UNET+Gaze \cite{karargyris2021creation} & \text{0.840$\pm$0.011} & \textbf{0.868$\pm$0.005} \\

\bottomrule
\end{tabular}
\end{table}

%% file: latex/tables/smaller-models-v3.tex
\begin{table*}[!htb]
\centering
\scriptsize
\caption{Ablation study with different sized ResNets for all tasks. AUC Means and standard deviations are reported across 10 independent runs, as well as the number of trainable parameters (M). The best results for each backbone are \textbf{bolded}, second best are \underline{underlined}.}
\label{tab:resnet-ablation-study}
\begin{tabular}{lllll}
\toprule
\textbf{Backbone} & \textbf{Params (M)} & \textbf{Iris PAD} & \textbf{Face} & \textbf{Chest} \\
\midrule
\textbf{ResNet18} & \text{} & \text{} & \text{} & \textbf{} \\
\hspace{2mm}Cross-entropy & 11.7 & \text{0.862$\pm$0.005} & \text{0.475$\pm$0.038} & \text{0.834$\pm$0.004} \\
\hspace{2mm}CYBORG \cite{boyd2021cyborg} & 11.7 & \textbf{0.890$\pm$0.019} & 0.614$\pm$0.028 & 0.834$\pm$0.010 \\
\hspace{2mm}UNET+Gaze \cite{karargyris2021creation} & 14.3 & \text{0.818$\pm$0.064} & 0.412$\pm$0.053 & 0.819$\pm$0.012 \\
\hspace{2mm}MENTOR (ours) & 11.7 & \underline{0.875$\pm$0.012} & 0.450$\pm$0.038 & \textbf{0.847$\pm$0.006} \\
\midrule
\textbf{ResNet34} & \textbf{} & \textbf{} & \textbf{} \\
\hspace{2mm}Cross-entropy & 21.8 & \text{0.887$\pm$0.020} & \text{0.510$\pm$0.057} & \text{0.823$\pm$0.017} \\
\hspace{2mm}CYBORG \cite{boyd2021cyborg} & 21.8 & \textbf{0.894$\pm$0.017} & 0.559$\pm$0.055 & \textbf{0.841$\pm$0.007} \\
\hspace{2mm}UNET+Gaze \cite{karargyris2021creation} & 24.4 & \text{0.789$\pm$0.097} & 0.399$\pm$0.049 & 0.815$\pm$0.008 \\
\hspace{2mm}MENTOR (ours) & 21.8 & \underline{0.893$\pm$0.015} & 0.534$\pm$0.053 & \underline{0.837$\pm$0.01} \\
\midrule
\textbf{ResNet50} & \textbf{} & \textbf{} & \textbf{} & \textbf{} \\
\hspace{2mm}Cross-entropy & 25.6 & \text{0.874$\pm$0.017} & 0.500$\pm$0.055 & 0.833$\pm$0.011 \\
\hspace{2mm}CYBORG \cite{boyd2021cyborg} & 25.6 & \text{0.875$\pm$0.028} & \textbf{0.531$\pm$0.050} & \textbf{0.846$\pm$0.010} \\
\hspace{2mm}UNET+Gaze \cite{karargyris2021creation} & 32.5 & \text{0.879$\pm$0.022} & 0.411$\pm$0.040 & 0.842$\pm$0.020 \\
\hspace{2mm}MENTOR (ours) & 25.6 & \textbf{0.912$\pm$0.028} & \underline{0.506$\pm$0.044} & \underline{0.844$\pm$0.006} \\
\midrule
\textbf{ResNet152} & \textbf{} & \textbf{} & \textbf{} & \textbf{ } \\
\hspace{2mm}Cross-entropy & 60.2 & \text{0.879$\pm$0.02} & \text{0.540$\pm$0.04} & \text{0.840$\pm$0.01} \\
\hspace{2mm}CYBORG \cite{boyd2021cyborg} & 60.2 & \text{0.897$\pm$0.02} & 0.508$\pm$0.05 & 0.854$\pm$0.01 \\
\hspace{2mm}UNET+Gaze \cite{karargyris2021creation} & 67.2 & \text{0.892$\pm$0.03} & 0.432$\pm$0.042 & 0.840$\pm$0.011 \\
\hspace{2mm}MENTOR (ours) & 60.2 & \textbf{0.912$\pm$0.02} & \textbf{0.538$\pm$0.05} & \textbf{0.856$\pm$0.01} \\
\bottomrule
\end{tabular}
\end{table*}

%% file: latex/tables/teacher-student-v3.tex
\begin{table*}
\centering
\scriptsize
\caption{Ablation study containing MENTOR's performance against the Teacher-Student paradigm for the \textbf{Iris PAD task}. For readability purposes, the table is reproduced following Crum \etal, and follows the same training configurations and evaluation measures. \textbf{Baseline 1} indicates small training set accompanied by human salience, \textbf{Baseline 2} indicates larger training sets \emph{without} human salience. The \textbf{Optimal AI Student \cite{crum2023teaching}} used a chain of models to provide CNN architectures with various forms of model saliency for larger training sets. MENTOR maintains competitive across half of the architecture without any additional fine-tuning. Best AUC is \textbf{bolded}, whereas second best AUC is \underline{underlined}.}
\label{tab:teacher-student}
\begin{tabular}{ccccc}
\toprule
\textbf{Model} & \textbf{Baseline 1 \cite{crum2023teaching}} & \textbf{Baseline 2 \cite{crum2023teaching}} & \textbf{Optimal AI Student \cite{crum2023teaching}} & \textbf{MENTOR (ours)} \\
\midrule
DenseNet121 & 0.920$\pm$0.017 & 0.917$\pm$0.017 & \textbf{0.950$\pm$0.013} & \underline{0.931$\pm$0.013} \\
ResNet50 & 0.854$\pm$0.031 & 0.905$\pm$0.013 & \underline{0.920$\pm$0.022} & \textbf{0.941$\pm$0.017} \\
Xception & 0.852$\pm$0.018 & 0.948$\pm$0.008 & \textbf{0.952$\pm$0.003} & \underline{0.947$\pm$0.004} \\
Inception-v3 & 0.888$\pm$0.018 & 0.905$\pm$0.029 & \underline{0.947$\pm$0.010} & \textbf{0.949$\pm$0.009} \\
\bottomrule
\end{tabular}
\end{table*}

%% file: latex/figures/saliency-sampler.tex
\begin{figure}[t]
  \centering
  \includegraphics[width=\linewidth]{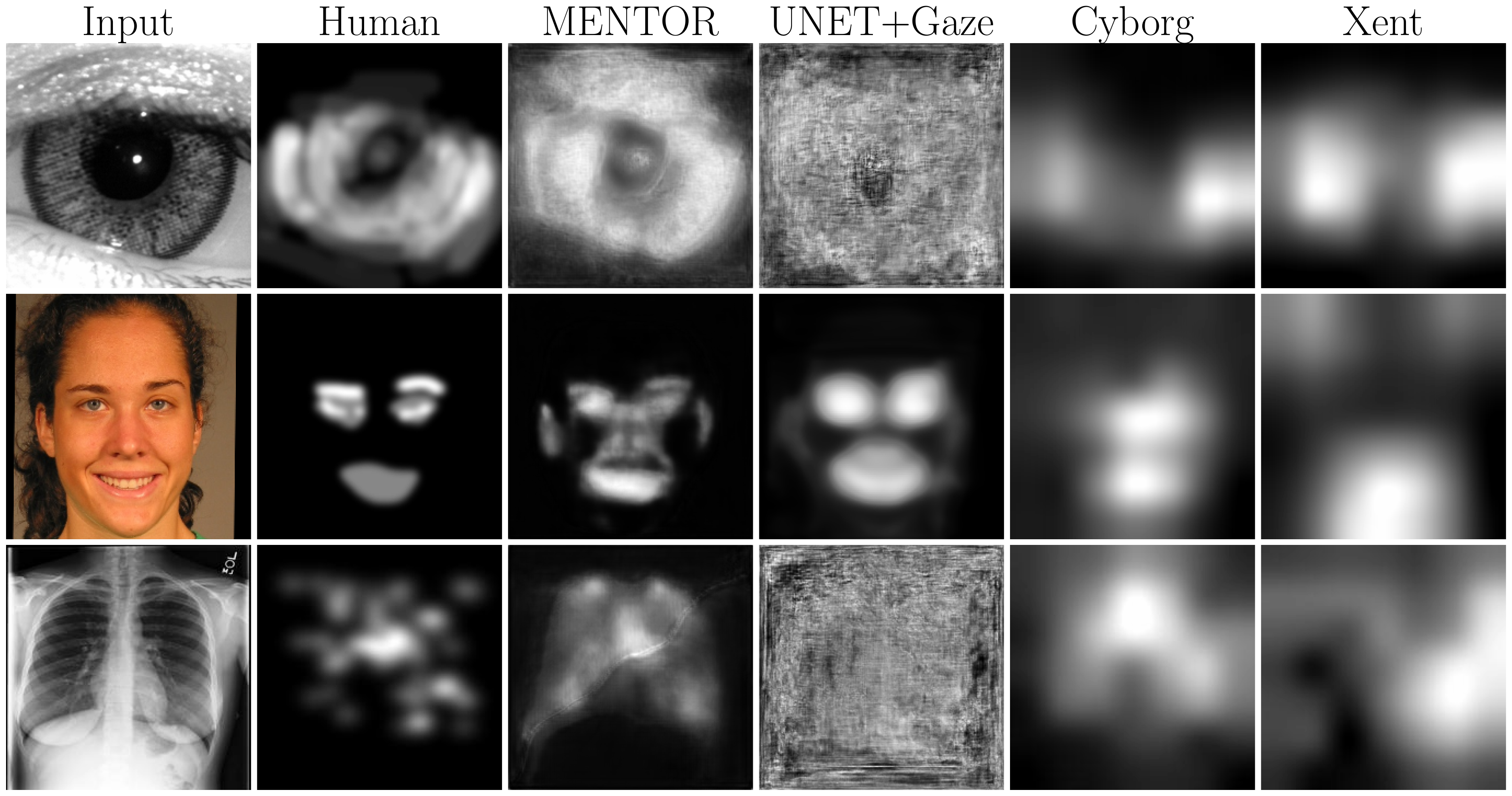}
  \caption{Same as Fig. \ref{fig:annotator}, except UNET+Gaze \cite{karargyris2021creation}, CYBORG \cite{boyd2021cyborg}, and Cross-entropy (Xent) saliency maps are shown. MENTOR and UNET+Gaze saliency are generated from their respective decoder, whereas Cyborg and Xent are generated using a Class Activation Mapping (CAM). Samples were generated from the validation set.}
  \label{fig:extended-annotator}
\end{figure}

%% file: latex/tables/ablation-entropy.tex
\begin{table}
\centering
\scriptsize
\caption{Ablation study of measuring the Salience Entropy ($S_{entropy}$) of the Class Activation Mapping (CAM) generated from the test set \cite{crum2023explain}. Means and standard deviations are reported for a \textbf{ResNet152} backbone across several different methods. Lower ($S_{entropy}$) typically suggest more compact or focused saliency.}
\label{tab:ablation-entropy}
\begin{tabular}{cccc}
\toprule
\textbf{Method} & \textbf{Iris PAD} & \textbf{Face} & \textbf{Chest} \\
\midrule
Cross-entropy & 0.958$\pm$0.018 & 0.959$\pm$0.017 & \text{0.960$\pm$0.016} \\
CYBORG \cite{boyd2021cyborg} & 0.963$\pm$0.016 & 0.942$\pm$0.021 & 0.938$\pm$0.023 \\
MENTOR (ours) & 0.961$\pm$0.017 & 0.959$\pm$0.017 & \text{0.960$\pm$0.018} \\
\bottomrule
\end{tabular}
\end{table}

%% file: latex/sections/06-conclusion-v1.tex
\section{Conclusion}
\label{sec:conclusion}

This paper introduces MENTOR, a novel and intuitively straightforward pretraining method that embeds the ``understanding'' of domain-specific human salient features into the model, different from previous human perception-guided training approaches, which align the spatial correspondence between human-sourced and model-sourced saliences. To do that, we first trained an autoencoder to learn associations between human-salient features and images of both classes. Once the latent representation between the input image and the human saliency is learned, we used the encoder part to build a domain-specific classifier. We showed that using MENTOR pretraining results in performance gains in classification tasks, even compared to a model whose weights were first transferred from a network trained on a vast number of images (ImageNet). We also demonstrated that MENTOR pretraining achieved better generalization than state-of-the-art approach to incorporate human salience during training via specially-designed loss function. We show that other human perceptual-guided training techniques initialized with MENTOR weights improves generalization performance when compared with traditional, non-human-perception-driven weights (ImageNet). Finally, we included a detailed ablation study highlighting the versatility of our method across the Teacher-Student Paradigm, salience generation, and various model sizes.

%% file: main.bbl
\begin{thebibliography}{10}\itemsep=-1pt

\bibitem{bigolin2022reflacx}
Ricardo Bigolin~Lanfredi, Mingyuan Zhang, William~F Auffermann, Jessica Chan, Phuong-Anh~T Duong, Vivek Srikumar, Trafton Drew, Joyce~D Schroeder, and Tolga Tasdizen.
\newblock Reflacx, a dataset of reports and eye-tracking data for localization of abnormalities in chest x-rays.
\newblock {\em Scientific data}, 9(1):350, 2022.

\bibitem{boyd2022human}
Aidan Boyd, Kevin~W Bowyer, and Adam Czajka.
\newblock Human-aided saliency maps improve generalization of deep learning.
\newblock In {\em Proceedings of the IEEE/CVF Winter Conference on Applications of Computer Vision}, pages 2735--2744, 2022.

\bibitem{boyd2023patch}
Aidan Boyd, Daniel Moreira, Andrey Kuehlkamp, Kevin Bowyer, and Adam Czajka.
\newblock Human saliency-driven patch-based matching for interpretable post-mortem iris recognition.
\newblock In {\em Proceedings of the IEEE/CVF Winter Conference on Applications of Computer Vision}, pages 701--710, 2023.

\bibitem{boyd2021cyborg}
Aidan Boyd, Patrick Tinsley, Kevin Bowyer, and Adam Czajka.
\newblock {CYBORG: Blending Human Saliency Into the Loss Improves Deep Learning-Based Synthetic Face Detection}.
\newblock In {\em {IEEE} Winter Conf. on Applications of Computer Vision (WACV)}, pages 6097--6106, 2023.

\bibitem{Caron_ICCV_2021}
Mathilde Caron, Hugo Touvron, Ishan Misra, Hervé Jegou, Julien Mairal, Piotr Bojanowski, and Armand Joulin.
\newblock Emerging properties in self-supervised vision transformers.
\newblock In {\em 2021 IEEE/CVF International Conference on Computer Vision (ICCV)}, pages 9630--9640, 2021.

\bibitem{casia-database}
Chinese academy of sciences institute of automation.
\newblock http://www.cbsr.ia.ac.cn/china/Iris\%20Databases\%20CH.asp.
\newblock Accessed: 03-12-2021.

\bibitem{stargan}
Yunjey Choi, Youngjung Uh, Jaejun Yoo, and Jung-Woo Ha.
\newblock Stargan v2: Diverse image synthesis for multiple domains.
\newblock In {\em Proceedings of the IEEE/CVF conference on computer vision and pattern recognition}, pages 8188--8197, 2020.

\bibitem{crum2023teaching}
Colton~R Crum, Aidan Boyd, Kevin Bowyer, and Adam Czajka.
\newblock Teaching ai to teach: Leveraging limited human salience data into unlimited saliency-based training.
\newblock {\em arXiv preprint arXiv:2306.05527}, 2023.

\bibitem{crum2023explain}
Colton~R Crum, Patrick Tinsley, Aidan Boyd, Jacob Piland, Christopher Sweet, Timothy Kelley, Kevin Bowyer, and Adam Czajka.
\newblock Explain to me: Salience-based explainability for synthetic face detection models.
\newblock {\em IEEE Transactions on Artificial Intelligence}, 2023.

\bibitem{czajka2019domain}
Adam Czajka, Daniel Moreira, Kevin Bowyer, and Patrick Flynn.
\newblock Domain-specific human-inspired binarized statistical image features for iris recognition.
\newblock In {\em 2019 IEEE Winter Conference on Applications of Computer Vision (WACV)}, pages 959--967. IEEE, 2019.

\bibitem{das2020iris}
Priyanka Das, Joseph McFiratht, Zhaoyuan Fang, Aidan Boyd, Ganghee Jang, Amir Mohammadi, Sandip Purnapatra, David Yambay, S{\'e}bastien Marcel, Mateusz Trokielewicz, et~al.
\newblock Iris liveness detection competition (livdet-iris)-the 2020 edition.
\newblock In {\em 2020 IEEE international joint conference on biometrics (IJCB)}, pages 1--9. IEEE, 2020.

\bibitem{Das_IJCB_2020}
P. {Das}, J. {Mcfiratht}, Z. {Fang}, A. {Boyd}, G. {Jang}, A. {Mohammadi}, S. {Purnapatra}, D. {Yambay}, S. {Marcel}, M. {Trokielewicz}, P. {Maciejewicz}, K. {Bowyer}, A. {Czajka}, S. {Schuckers}, J. {Tapia}, S. {Gonzalez}, M. {Fang}, N. {Damer}, F. {Boutros}, A. {Kuijper}, R. {Sharma}, C. {Chen}, and A. {Ross}.
\newblock {Iris Liveness Detection Competition (LivDet-Iris) - The 2020 Edition}.
\newblock In {\em 2020 IEEE International Joint Conference on Biometrics (IJCB)}, pages 1--9, 2020.

\bibitem{ImageNet}
Jia Deng, Wei Dong, Richard Socher, Li-Jia Li, Kai Li, and Li Fei-Fei.
\newblock Imagenet: A large-scale hierarchical image database.
\newblock In {\em 2009 IEEE Conference on Computer Vision and Pattern Recognition}, pages 248--255, 2009.

\bibitem{dulay2022using}
Justin Dulay and Walter~J Scheirer.
\newblock Using human perception to regularize transfer learning.
\newblock {\em arXiv preprint arXiv:2211.07885}, 2022.

\bibitem{Galbally_ICB_2012}
Javier Galbally, Jaime Ortiz-Lopez, Julian Fierrez, and Javier Ortega-Garcia.
\newblock Iris liveness detection based on quality related features.
\newblock In {\em 2012 5th IAPR Int. Conf. on Biometrics (ICB)}, pages 271--276, New Delhi, India, March 2012. IEEE.

\bibitem{gopnik1999scientist}
Alison Gopnik, Andrew~N Meltzoff, and Patricia~K Kuhl.
\newblock {\em The scientist in the crib: Minds, brains, and how children learn.}
\newblock William Morrow \& Co, 1999.

\bibitem{he2016deep}
Kaiming He, Xiangyu Zhang, Shaoqing Ren, and Jian Sun.
\newblock Deep residual learning for image recognition.
\newblock In {\em Proceedings of the IEEE conference on computer vision and pattern recognition}, pages 770--778, 2016.

\bibitem{huang2023measuring}
Jin Huang, Derek Prijatelj, Justin Dulay, and Walter Scheirer.
\newblock Measuring human perception to improve open set recognition.
\newblock {\em IEEE Transactions on Pattern Analysis and Machine Intelligence}, 2023.

\bibitem{Iakubovskii:2019}
Pavel Iakubovskii.
\newblock Segmentation models pytorch.
\newblock \url{https://github.com/qubvel/segmentation_models.pytorch}, 2019.

\bibitem{Ismail_NeurIPS_2021}
Aya~Abdelsalam Ismail, H{\'e}ctor~Corrada Bravo, and Soheil Feizi.
\newblock Improving deep learning interpretability by saliency guided training.
\newblock In {\em Neural Information Processing Systems (NeurIPS)}, 2021.

\bibitem{johnson2019mimic}
Alistair~EW Johnson, Tom~J Pollard, Seth~J Berkowitz, Nathaniel~R Greenbaum, Matthew~P Lungren, Chih-ying Deng, Roger~G Mark, and Steven Horng.
\newblock Mimic-cxr, a de-identified publicly available database of chest radiographs with free-text reports.
\newblock {\em Scientific data}, 6(1):317, 2019.

\bibitem{karargyris2021creation}
Alexandros Karargyris, Satyananda Kashyap, Ismini Lourentzou, Joy~T Wu, Arjun Sharma, Matthew Tong, Shafiq Abedin, David Beymer, Vandana Mukherjee, Elizabeth~A Krupinski, et~al.
\newblock Creation and validation of a chest x-ray dataset with eye-tracking and report dictation for ai development.
\newblock {\em Scientific data}, 8(1):92, 2021.

\bibitem{karras2017progressive}
Tero Karras, Timo Aila, Samuli Laine, and Jaakko Lehtinen.
\newblock {Progressive Growing of GANs for Improved Quality, Stability, and Variation}.
\newblock {\em arXiv preprint arXiv:1710.10196}, 2017.

\bibitem{styleGAN}
Tero Karras, Timo Aila, Samuli Laine, and Jaakko Lehtinen.
\newblock Progressive growing of gans for improved quality, stability, and variation.
\newblock {\em arXiv preprint arXiv:1710.10196}, 2017.

\bibitem{Celeb}
Tero Karras, Timo Aila, Samuli Laine, and Jaakko Lehtinen.
\newblock Progressive growing of gans for improved quality, stability, and variation.
\newblock {\em arXiv preprint arXiv:1710.10196}, 2017.

\bibitem{syleGAN2-ADA}
Tero Karras, Miika Aittala, Janne Hellsten, Samuli Laine, Jaakko Lehtinen, and Timo Aila.
\newblock Training generative adversarial networks with limited data.
\newblock {\em Advances in neural information processing systems}, 33:12104--12114, 2020.

\bibitem{styleGAN3}
Tero Karras, Miika Aittala, Samuli Laine, Erik H{\"a}rk{\"o}nen, Janne Hellsten, Jaakko Lehtinen, and Timo Aila.
\newblock Alias-free generative adversarial networks.
\newblock {\em Advances in Neural Information Processing Systems}, 34:852--863, 2021.

\bibitem{FFHQ}
Tero Karras, Samuli Laine, and Timo Aila.
\newblock A style-based generator architecture for generative adversarial networks.
\newblock In {\em IEEE/CVF conference on computer vision and pattern recognition}, pages 4401--4410, 2019.

\bibitem{styleGAN2}
Tero Karras, Samuli Laine, Miika Aittala, Janne Hellsten, Jaakko Lehtinen, and Timo Aila.
\newblock Analyzing and improving the image quality of stylegan.
\newblock In {\em Proceedings of the IEEE/CVF conference on computer vision and pattern recognition}, pages 8110--8119, 2020.

\bibitem{Kohli_ICB_2013}
Naman Kohli, Daksha Yadav, Mayank Vatsa, and Richa Singh.
\newblock Revisiting iris recognition with color cosmetic contact lenses.
\newblock In {\em {IEEE} Int. Conf. on Biometrics (ICB)}, pages 1--7, Madrid, Spain, June 2013. IEEE.

\bibitem{Kohli_BTAS_2016}
Naman Kohli, Daksha Yadav, Mayank Vatsa, Richa Singh, and Afzel Noore.
\newblock Detecting medley of iris spoofing attacks using desist.
\newblock In {\em {IEEE} Int. Conf. on Biometrics: Theory Applications and Systems (BTAS)}, pages 1--6, Niagara Falls, NY, USA, Sept 2016. IEEE.

\bibitem{Sung_OE_2007}
Sung~Joo Lee, Kang~Ryoung Park, Youn~Joo Lee, Kwanghyuk Bae, and Jai~Hie Kim.
\newblock {Multifeature-based fake iris detection method}.
\newblock {\em Optical Engineering}, 46(12):1 -- 10, 2007.

\bibitem{linsley2018learning}
Drew Linsley, Dan Shiebler, Sven Eberhardt, and Thomas Serre.
\newblock Learning what and where to attend.
\newblock {\em arXiv preprint arXiv:1805.08819}, 2018.

\bibitem{parzianello2022saliency}
Lucas Parzianello and Adam Czajka.
\newblock Saliency-guided textured contact lens-aware iris recognition.
\newblock In {\em Proceedings of the IEEE/CVF Winter Conference on Applications of Computer Vision Workshops (WACVW)}, pages 330--337, 2022.

\bibitem{petsiuk2018rise}
Vitali Petsiuk, Abir Das, and Kate Saenko.
\newblock Rise: Randomized input sampling for explanation of black-box models.
\newblock {\em arXiv preprint arXiv:1806.07421}, 2018.

\bibitem{Pham_WACV_2024}
Trong~Thang Pham, Jacob Brecheisen, Anh Nguyen, Hien Nguyen, and Ngan Le.
\newblock {I-AI: A Controllable \& Interpretable AI System for Decoding Radiologists’ Intense Focus for Accurate CXR Diagnoses}.
\newblock In {\em {IEEE} Winter Conf. on Applications of Computer Vision (WACV)}, pages 7850--7859, Waikoloa, HI, USA, January 2024. IEEE.

\bibitem{piland2023model}
Jacob Piland, Adam Czajka, and Christopher Sweet.
\newblock Model focus improves performance of deep learning-based synthetic face detectors.
\newblock {\em IEEE Access}, 2023.

\bibitem{Rombach_CVPR_2022}
Robin Rombach, Andreas Blattmann, Dominik Lorenz, Patrick Esser, and Björn Ommer.
\newblock High-resolution image synthesis with latent diffusion models.
\newblock In {\em 2022 IEEE/CVF Conference on Computer Vision and Pattern Recognition (CVPR)}, pages 10674--10685, 2022.

\bibitem{ronneberger2015u}
Olaf Ronneberger, Philipp Fischer, and Thomas Brox.
\newblock U-net: Convolutional networks for biomedical image segmentation.
\newblock In {\em Medical Image Computing and Computer-Assisted Intervention--MICCAI 2015: 18th International Conference, Munich, Germany, October 5-9, 2015, Proceedings, Part III 18}, pages 234--241. Springer, 2015.

\bibitem{szegedy2017inception}
Christian Szegedy, Sergey Ioffe, Vincent Vanhoucke, and Alexander Alemi.
\newblock Inception-v4, inception-resnet and the impact of residual connections on learning.
\newblock In {\em Proceedings of the AAAI conference on artificial intelligence}, volume~31, 2017.

\bibitem{tan2019efficientnet}
Mingxing Tan and Quoc Le.
\newblock Efficientnet: Rethinking model scaling for convolutional neural networks.
\newblock In {\em International conference on machine learning}, pages 6105--6114. PMLR, 2019.

\bibitem{Trokielewicz_BTAS_2015}
M. {Trokielewicz}, A. {Czajka}, and P. {Maciejewicz}.
\newblock Assessment of iris recognition reliability for eyes affected by ocular pathologies.
\newblock In {\em {IEEE} Int. Conf. on Biometrics: Theory Applications and Systems (BTAS)}, pages 1--6, 2015.

\bibitem{Trokielewicz_IVC_2020}
Mateusz Trokielewicz, Adam Czajka, and Piotr Maciejewicz.
\newblock Post-mortem iris recognition with deep-learning-based image segmentation.
\newblock {\em Image and Vision Computing}, 94:103866, 2020.

\bibitem{van2023probabilistic}
Tom van Sonsbeek, Xiantong Zhen, Dwarikanath Mahapatra, and Marcel Worring.
\newblock Probabilistic integration of object level annotations in chest x-ray classification.
\newblock In {\em Proceedings of the IEEE/CVF Winter Conference on Applications of Computer Vision}, pages 3630--3640, 2023.

\bibitem{Wei_ICPR_2008}
Zhuoshi Wei, Tieniu Tan, and Zhenan Sun.
\newblock Synthesis of large realistic iris databases using patch-based sampling.
\newblock In {\em Int. Conf. on Pattern Recognition (ICPR)}, pages 1--4, Tampa, FL, USA, Dec 2008. IEEE.

\bibitem{Yambay_IJCB_2017}
David Yambay, Benedict Becker, Naman Kohli, Daksha Yadav, Adam Czajka, Kevin~W. Bowyer, Stephanie Schuckers, Richa Singh, Mayank Vatsa, Afzel Noore, Diego Gragnaniello, C. Sansone, L. Verdoliva, Lingxiao He, Yiwei Ru, Haiqing Li, Nianfeng Liu, Zhenan Sun, and Tieniu Tan.
\newblock {LivDet Iris 2017 -- Iris Liveness Detection Competition 2017}.
\newblock In {\em {IEEE} Int. Joint Conf. on Biometrics (IJCB)}, pages 1--6, Denver, CO, USA, 2017. IEEE.

\bibitem{Yambay_ISBA_2017}
David Yambay, Brian Walczak, Stephanie Schuckers, and Adam Czajka.
\newblock {LivDet-Iris 2015 - Iris Liveness Detection Competition 2015}.
\newblock In {\em {IEEE} Int. Conf. on Identity, Security and Behavior Analysis (ISBA)}, pages 1--6, New Delhi, India, Feb 2017. IEEE.

\bibitem{zhou2016learning}
Bolei Zhou, Aditya Khosla, Agata Lapedriza, Aude Oliva, and Antonio Torralba.
\newblock Learning deep features for discriminative localization.
\newblock In {\em Proceedings of the IEEE conference on computer vision and pattern recognition}, pages 2921--2929, 2016.

\bibitem{zhou2018unet++}
Zongwei Zhou, Md~Mahfuzur Rahman~Siddiquee, Nima Tajbakhsh, and Jianming Liang.
\newblock Unet++: A nested u-net architecture for medical image segmentation.
\newblock In {\em Deep Learning in Medical Image Analysis and Multimodal Learning for Clinical Decision Support: 4th International Workshop, DLMIA 2018, and 8th International Workshop, ML-CDS 2018, Held in Conjunction with MICCAI 2018, Granada, Spain, September 20, 2018, Proceedings 4}, pages 3--11. Springer, 2018.

\end{thebibliography}
